\documentclass[letterpaper, 10 pt, conference]{ieeeconf}  

\IEEEoverridecommandlockouts                              
\usepackage[utf8]{inputenc}
\usepackage[T1]{fontenc}
\usepackage{graphicx}
\usepackage{multirow}

\title{\LARGE \bf Third-party Evaluation of Robotic Hand Designs \\
Using a Mechanical Glove}

\author{Takayuki Kanai$^{1}$, Yoshiyuki Ohmura$^{1}$, Akihiko Nagakubo$^{2}$, and Yasuo Kuniyoshi$^{1}$
\thanks{$^{1}$T. Kanai, Y. Ohmura, and Y. Kuniyoshi are with Laboratory for Intelligent Systems and Informatics, Graduate School of Information Science and Technology, University of Tokyo, 7-3-1 Hongo, Bunkyo-ku, Tokyo, Japan, e-mail:
{\tt\small \{kanai,ohmura,kuniyosh\}
@isi.imi.i.u-tokyo.ac.jp}}
\thanks{$^{2}$A. Nagakubo is with Artificial Intelligence Research Center, National Institute of Advanced Industrial Science and Technology, 1-1-1 Umezono, Tsukuba, Ibaraki ,Japan, e-mail:
{\tt\small nagakubo.a@aist.go.jp}}
}

\begin{document}
\bstctlcite{IEEEexample:BSTcontrol}

\maketitle
\thispagestyle{empty}
\pagestyle{empty}

\begin{abstract}
A robotic hand design suitable for dexterity should be examined using functional tests. To achieve this, we designed a mechanical glove, which is a rigid wearable glove that enables us to develop the corresponding isomorphic robotic hand and evaluate its hardware properties. Subsequently, the effectiveness of multiple degrees-of-freedom (DOFs) was evaluated by human participants. Several fine motor skills were evaluated using the mechanical glove under two conditions: one- and three-DOF conditions. To the best of our knowledge, this is the first extensive evaluation method for robotic hand designs suitable for dexterity.
(This paper was peer-reviewed and is the full translation from the Journal of the Robotics Society of Japan, Vol.39, No.6, pp.557-560, 2021.)

\end{abstract}


\section{Introduction}

How do embodiments contribtute to dexterity? In this study, dexterity is not defined as the ability to specialize in a specific task; however, it is defined as the ability to adapt to various objects and environments. Dexterity is determined by both physical characteristics and the whole system, including intelligence. Therefore, robotic hands must not be evaluated only by physical properties, such as possible hand postures or the maximal force applied to an object \cite{Palli2014,Santia2018}, but through the performance of various functional tests \cite{HuamanQuispe2018}. Considering such functional tests, it is difficult to separate the contributions of physical characteristics and intelligence because they strongly interact with each other to perform tasks. To clarify the contributions of the physical characteristics, it is necessary to quantitatively evaluate the changes in behavior when only the physical characteristics are changed in a dexterous system.

To appropriately evaluate the physical characteristics through various tests, a strategy that maximizes the performance of a specific hardware should be chosen for each condition. However, the operation strategy and adaptability of robotic hands are currently far from human hand skills \cite{Ozawa2017}. To achieve maximal performance, we assume that the direct operation of robotic hands by humans is the most promising method. Considering objectivity, third-party evaluation is also important. 


In this study, we propose a novel wearable mechanical glove that can be used by a third party to evaluate robotic hand designs. Furthermore, we evaluate the robotic hand design by comparing the results of manipulation experiments conducted by a third party using mechanical gloves under two different conditions. This evaluation method has a low barrier to entry because it does not require high technology for robot implementation and is suitable for large-scale competitions.

\begin{figure}[t]
\begin{center}
\includegraphics[width=80mm]{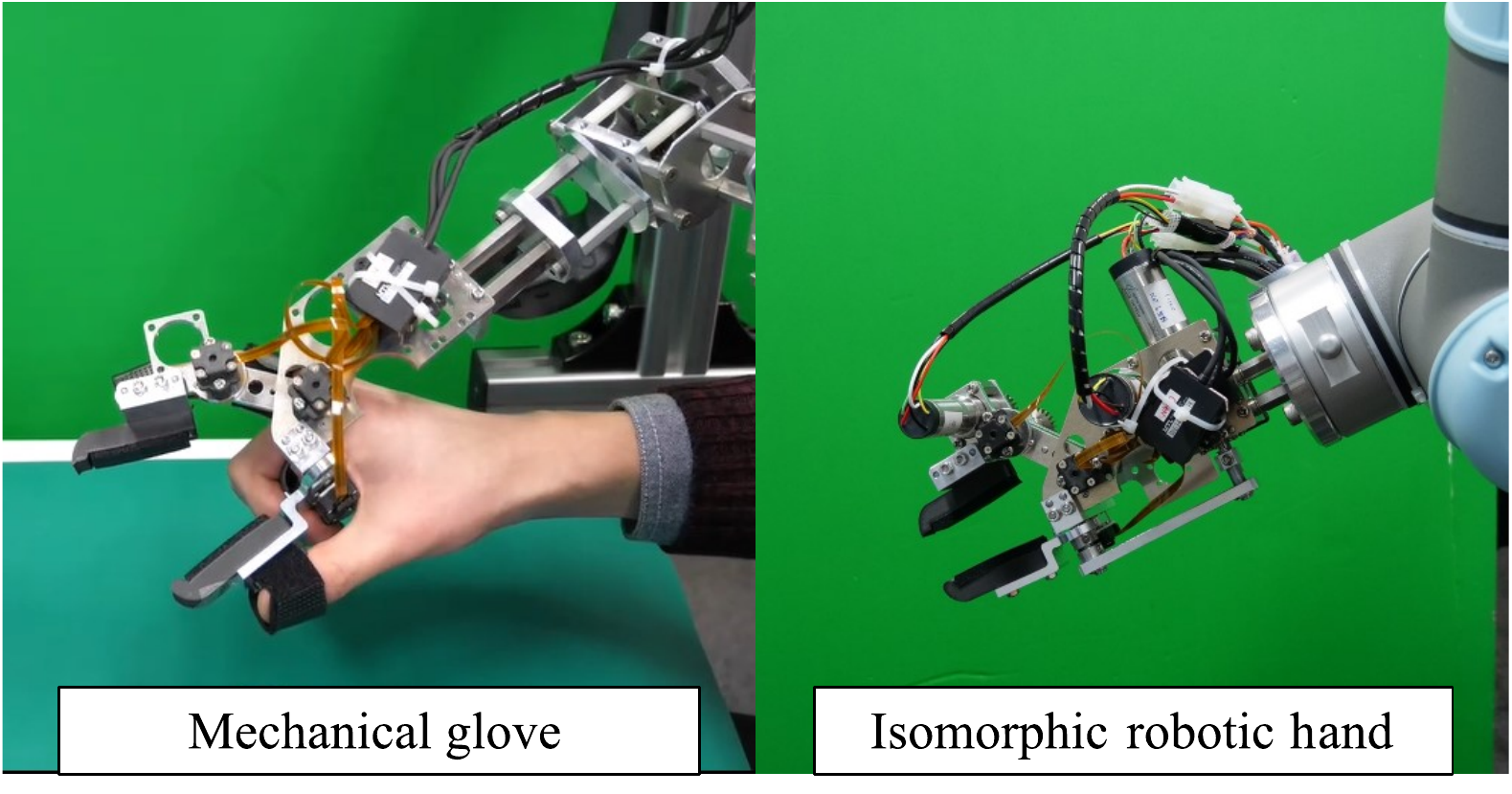}
	\end{center}
	\caption{Overview of the developed robot}
	\label{fig:developed_hand}
\end{figure}


\section{Third-party Evaluation Method \\ of the Robotic Hand Design}
\subsection{Mechanical Glove and Isomorphic Robotic Hand}
A mechanical glove satisfies the following three conditions.
\begin{itemize}
    \item An isomorphic robotic hand to compare the mechanical design can be built.
    \item The body is rigid enough to keep the shape.
    \item Third-party participants can wear the same mechanical glove and handle objects easily.
\end{itemize}
The most important feature of the proposed mechanical glove is that it is composed of a rigid joint link system which enables the construction of an isomorphic robotic hand. Hereinafter, a pair of mechanical gloves and isomorphic robotic hands are defined as isomorphic systems. Although conventional wearable sensing gloves are often made of soft materials for easy use, controlling their experimental conditions is difficult because of their low rigidity and inaccurate joint link model. We assume that this softness hinders the elucidation of the mechanical properties for dexterity. On the contrary, mechanical gloves with rigid joint links enable the comparison of mechanical designs under controlled conditions. 

A mechanical glove requires sufficient clearance between the user hand and mechanical structures for usability. If a human hand is tightly fixed with the glove, a closed link structure between the hand and glove strongly restricts hand movement.

To build an isomorphic system, the robotic hand design must be limited by its usability for humans. Therefore, the contribution of the mechanical properties cannot be evaluated independently from the shape of the human hand. However, this drawback does not arise only with our method. The same problem arises in teleoperation if the operating controller and robot have an isomorphic structure. If the heteromorphic structure, a mapping problem \cite{Yang2017AMC} that translates human motions to a robot must be solved. Therefore, separating usability and dexterity in a human-in-the-loop evaluation is difficult regardless of an teleoperation manner. Another limitation is the loss of tactile sensation owing to the rigidity of the glove. In particular, tactile texture information is hindered and the contribution of visual information increases in our method. In contrast, compared to a remote control, the wearable type can realize force control in a more natural way without considering that the performance of the bilateral control will affect the evaluation results. 

\subsection{Skill Evaluation through Fine Motor Skill}
The evaluation result strongly depends on the type of manipulation test.
Considering conventional robotic manipulation studies, many
benchmark tasks for manipulation have been proposed \cite{BerkCALLI2017,Young2009}; nonetheless, they are inappropriate for our purpose because they are set as tasks for an automatic robot and not intended to investigate the physical properties contributing to dexterity. Therefore, many kinds of tasks are suitable for evaluation because dexterity is an adaptation ability in a wide variety of situations. However, there are various skills from daily activities to craftsmanship, and the variety makes it unrealistic to investigate comprehensively in subject experiments with limited time. Therefore, we consider setting a standard that correlates with the required physical and cognitive skills to quantify the level of the target skills and ignore the slight differences between tasks. 

Consequently, in this study, we used the developmental acquisition age of fine motor skills \cite{Gallahue2012} as a measure of skill level. Infants tend to acquire important skills more rapidly than rarely used skills \cite{nla_cat_vn392109}, and skills learned first do not require skills learned later. Moreover, the required cognitive skills become more sophisticated with age. Therefore, we assume that the evaluation tasks can be randomly selected from the fine motor skills developed by the target age.


\subsection{Evaluation Flow of Robotic Hand Design}
A golden standard method of evaluating various designs is required to clarify the contribution of the physical properties to dexterity. In many conventional studies, the evaluation of robot design has not been investigated quantitatively by a third party. In our method, a third-party participant conducts several fine motor skills using a mechanical glove under several conditions to evaluate the effect of the target mechanical properties. After the experiments, the mechanical properties that are crucial for the target skill level can be clarified using a statistical comparison test of the performance.

\section{Implementation}
We developed a mechanical glove and corresponding isomorphic robotic hand. First, we examined the size of the available actuator. The maximum fingertip force required for the manipulation of daily life objects ranged from 30 to 50 N in our preliminary experiments. We selected a brushless direct current (BLDC) motor (ECX Series, Maxon Motor Inc.) to meet the above requirements.

\begin{figure}[t]
    \begin{center}
    \includegraphics[width=80mm]{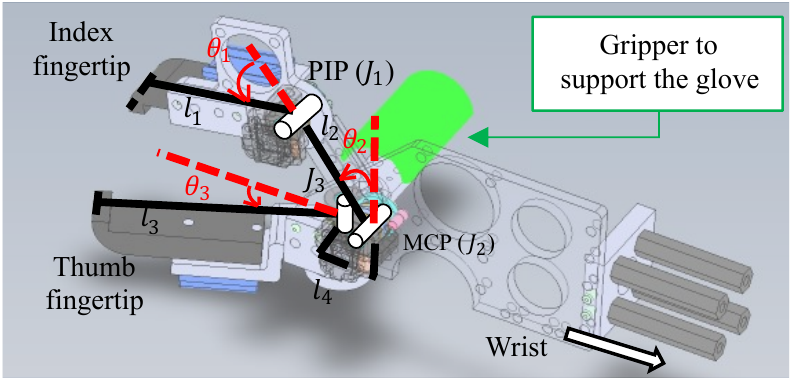}
    \end{center}
        \caption{Kinematics of the mechanical glove}
        \label{fig:kinematics}
\end{figure}

\begin{figure}[b]
    \begin{center}
    \includegraphics[height=30mm]{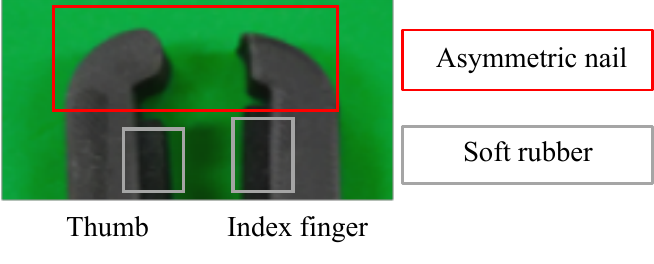} 
    \end{center}
        \caption{Details of the fingertips}
        \label{fig:nail}
\end{figure}

Subsequently, we designed the mechanical structures of the isomorphic system. We used gears and link mechanisms as the transmission system without wires to realize the precise motion control and joint rigidity although it limits the possible actuator placements.

The implemented mechanical glove and isomorphic robotic hand are shown in Fig.\ref{fig:developed_hand}; the definitions of the parameters and joint names are shown in Fig.\ref{fig:kinematics}; the link lengths and movable joint angles are shown in Table \ref{tab:link} and Table \ref{tab:angle}.  To minimize the variations in the robot design, we designed a three-DOF robotic hand with a two-DOF index finger (corresponding to the proximal interphalangeal (PIP) joint and metacarpophalangeal (MCP) joint of the finger) and one-DOF thumb. Other mechanical components (such as link parameters, range of motion, shapes, and materials) were also optimized using a rapid prototyping process. In particular, the finger essential for interaction with objects and the environment was composed of asymmetrical nails and covered with a soft material sheet (Fig.\ref{fig:nail}). All angles of the joints were measured using absolute rotary encoders (MAS-3-4096N1, Microtech Laboratory Inc.). 

\begin{table}[h]
    \begin{center}
    \begin{tabular}{cc}
    \begin{minipage}{0.45\hsize}
        \begin{center}
        \caption{Link length}
        \label{tab:link}
        \begin{tabular}{|c|c|}
            \hline
            $l1$ & 67.6 mm \\ \hline
            $l2$ & 40.0 mm \\ \hline
            $l3$ & 73.2 mm \\ \hline
            $l4$ & 17.0 mm \\ \hline
        \end{tabular}
        \end{center}
    \end{minipage}

    \begin{minipage}{0.45\hsize}
        \begin{center}
        \caption{Joint angles}
        \label{tab:angle}
        \begin{tabular}{|c|c|}
            \hline
            $\theta_{1}$ & [ 0$^\circ$, 99$^\circ$] \\ \hline
            $\theta_{2}$ & [ 0$^\circ$,  90$^\circ$] \\ \hline
            $\theta_{3}$ & [ -46$^\circ$, 53$^\circ$] \\ \hline
        \end{tabular}
        \end{center}
    \end{minipage}
    \end{tabular}
    \end{center}
\end{table}


The details of the transmission mechanisms of the robotic hand are as follows. Joint $J_{1}$ is driven by the gear (gear ratio of 1:1), and $J_{2}$ and $J_{3}$ are driven by parallel link mechanisms. The maximum force of the fingertip is 30 N. To realize this, we used 40 W, 80 W, and 80 W motors at $(J_{1}, J_{2}, J_{3})$, whose reduction ratios were 326, 231, and 231, respectively. 


Considering a mechanical glove, realizing high usability and reducing the effects of variability in the user’s hand size is important. The user’s thumb and index finger are loosely fixed to the links of the mechanical glove with Velcro to realize it. However, loose fixation makes it difficult to stabilize the posture of the glove. To resolve this, the stabilized gripper is attached to the glove that the user can hold with the middle finger during object manipulation. 

\section{Experimental Evaluation}
\subsection{Evaluation of the Degree of Freedom}
A large degree of freedom seems suitable for stable grasping and object manipulation. In particular, object manipulation is required for in-hand manipulations. However, humans often slip objects in their hands or find a robust strategy against physical constraints. Considering the human-level handling strategy, a large degree of freedom of the finger may not always contribute to the object manipulation skill. To answer this question, in this study, we experimentally examined the contribution of multiple DOFs to task performance. 

\begin{figure}[b]
\begin{center}
\includegraphics[width=80mm]{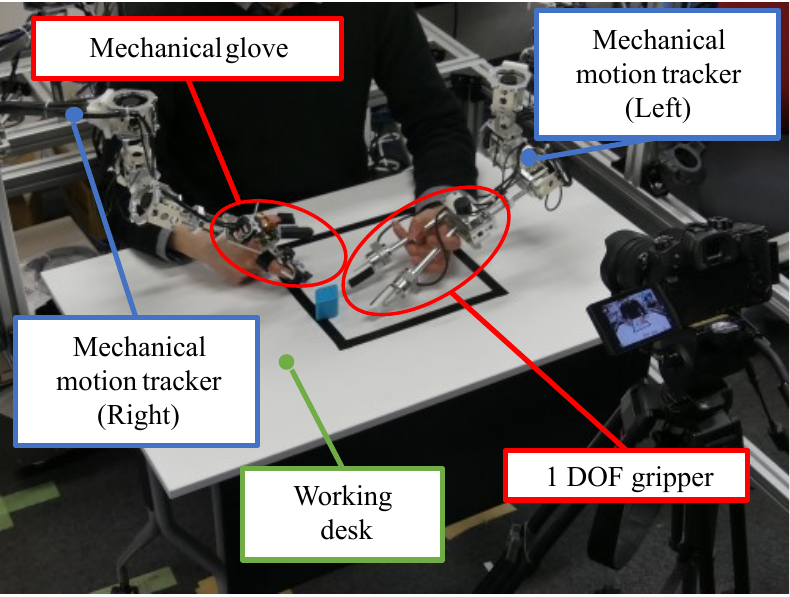}
	\end{center}
	\caption{Overview of the experimental setup}
	\label{fig:eval_setup}
\end{figure}

\subsection{Experimental Setup}
This study was approved by the research ethics committee of the University of Tokyo. The experiments were conducted in accordance with the relevant guidelines and regulations based on the Declaration of Helsinki (1964). All the participants provided written informed consents. After the experiment, the participants received monetary compensation for their time.

In the experiment, 14 people with a mean age of 23 years (standard deviation = 2.6, three females) participated in the experiment and completed the tasks. All the participants were right-handed, Japanese speakers, and were able to put on the glove.

Fig.\ref{fig:eval_setup} shows the overview of the experimental setup. The mechanical glove is attached to the right arm of the dual-arm mechanical motion tracker which can collect joint angle data using absolute rotary encoders (MAS-18-262144N1, Microtech Laboratory Inc.). The mechanical motion tracker also has the same link joint model as the six-axis arm robot, UR5. Therefore, participants were only allowed to move in the range the actual dual-arm robot can realize. Participants conducted tasks using the mechanical glove on the right hand and one-DOF gripper on the left hand, if required. Therefore, all the participants shared the same embodiment with the robot in terms of the movable range.

\begin{figure}[t]
    \begin{center}
    \includegraphics[width=68mm]{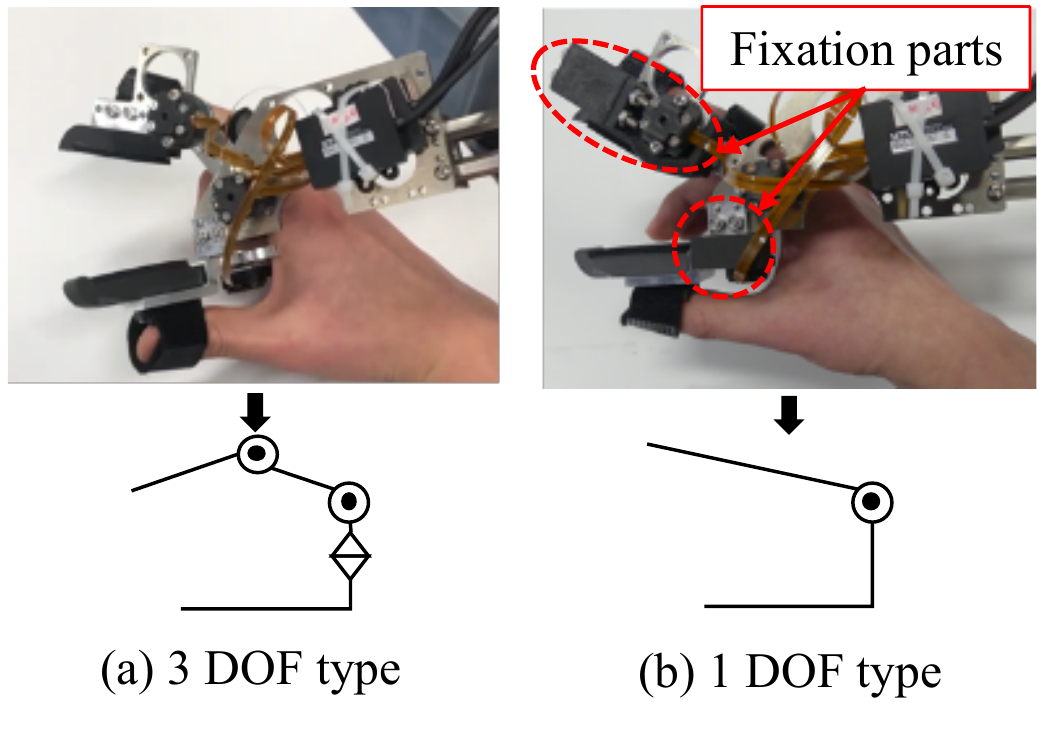}
        \end{center}
        \caption{Two conditions for using the mechanical glove: (a) original three-degree-of-freedom (DOF) model and (b) restricted one-DOF model.}
        \label{fig:two_conditions}
\end{figure}

The participants conducted the tasks using a mechanical glove under two conditions (Fig.\ref{fig:two_conditions}): three- (Fig.\ref{fig:two_conditions}-a) and one-DOF conditions with two joints fixed by custom-made jigs (Fig.\ref{fig:two_conditions}-b). The length of the fingertip was shortened according to the change in the DOF such that the position of the fingertips could be aligned.

\subsection{Evaluation Tasks}
\begin{figure*}[t]
\begin{center}
\includegraphics[width=16cm]{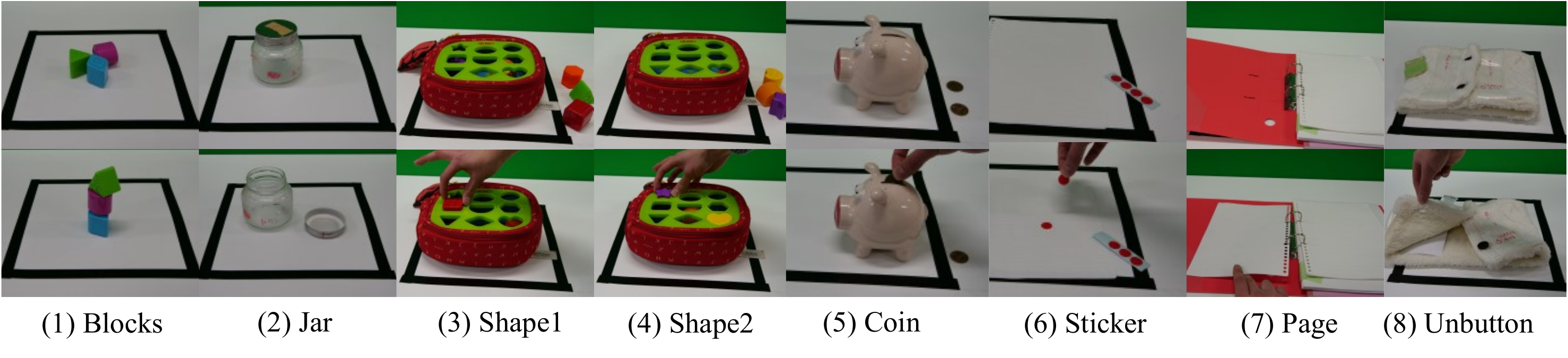}
	\end{center}
	\caption{Fine motor skill-based evaluation tasks used in the experiment (top: initial state; bottom: end state)}
	\label{fig:task_list_highres}
\end{figure*}

The evaluation tasks were chosen based on the fine motor skills of toddlers and preschool children (Fig. \ref{fig:task_list_highres}): (1) building a tower using toy blocks (including triangle, rhombus, and elliptical shape); (2) opening the lid of a jar; (3) shape sorting using three types of blocks (triangle, square, and hexagonal); (4) shape sorting using another set of blocks (semicircle, star, and heart); (5) inserting a coin in a piggy bank; (6) putting a sticker on a sheet of paper; (7) turning a single page of a loose-leaf file; (8) unbuttoning. The initial positions of the objects in each task were marked with a black taped line. Tasks (1) to (4), which require relatively gross motor skills, are assumed to be the skills of toddlers, and the others, which require relatively finer motor skills, are assumed to be the skills of preschool children \cite{nla_cat_vn392109}. To avoid bias, in the manipulation strategy, the subjects were not given prior instructions on accomplishing the tasks using the mechanical glove.

As a metric for measuring the performance of the robotic hand, we use the completion time of the tasks because time is one of the most important factors for evaluating the practicality of the robotic hands \cite{HuamanQuispe2018,Ortenzi2019}. Success rates are not used because all the tasks can be accomplished by adults. The tests were conducted chronologically (from one to eight) and repeated three times for each task. The experiment lasted for 1.5-2.0 h per person. To eliminate the ordering effects, we divided the participants into two groups: one group experienced the three-DOF condition before the one-DOF condition and the other group experienced the one-DOF condition before the three-DOF condition for all the tasks.

\subsection{Results}
First, we evaluated the contribution of DOF to the sum of the completion times of all the tasks. We calculated the total completion time of the eight tasks for each subject using the shortest time of three trials in each task. The results show that the DOF of the robotic hand does not have a significant effect on the completion time; nonetheless, the order of the DOF condition has a significant effect on the performance ($p<0.05$, interaction, two-way repeated ANOVA; the factors are DOF and the participant group, Table \ref{tab:total}).

\begin{figure}[b]
\begin{center}\includegraphics[width=8cm]{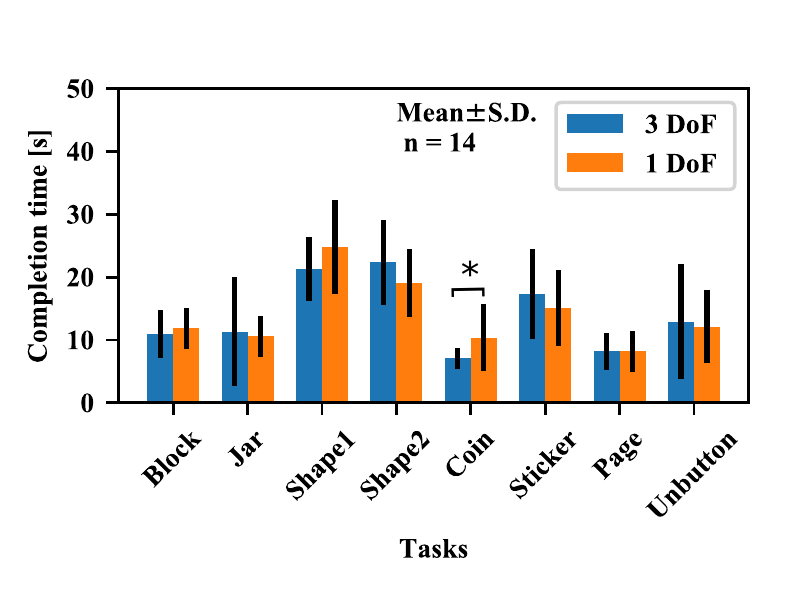}
	\end{center}
	\caption{Completion time of all the tasks ($p < 0.05$)}
	\label{fig:1_vs_3DoF_byouga}
\end{figure}

We examined tasks that could benefit from the DOF. Fig.\ref{fig:1_vs_3DoF_byouga} shows the shortest task-completion time among the three trials for each DOF condition. Considering the coin-insertion task, the working time for the three-DOF condition is significantly shorter than that for the one-DOF condition ($p < 0.05$, DOF, two-way repeated ANOVA, Table \ref{tab:coin}). However, no other significant differences in performance were observed in this study.

\begin{table}[ht]
\centering
\caption{Total working time at the best performance}
\label{tab:total}
\begin{tabular}{|c|c|c|}
\hline
\multirow{2}{*}{DOF} & \multicolumn{2}{c|}{Total task-completion time ( S.D. ) [s]}         \\ \cline{2-3} 
                              & 1 DOF first ($n=7$) & 3 DOFs first ($n=7$) \\ \hline
1 & 117.7 (37.8) & 106.6 (15.5) \\ \hline
3 & 99.8 (26.1)  & 123.0 (33.2) \\ \hline
\end{tabular}
\end{table}

\begin{table}[ht]
\centering
\caption{Working time on coin task at the best performance}
\label{tab:coin}
\begin{tabular}{|c|c|c|}
\hline
\multirow{2}{*}{DOF} & \multicolumn{2}{c|}{Total task-completion time ( S.D. ) [s]}         \\ \cline{2-3} 
                              & 1 DOF first ($n=7$) & 3 DOFs first ($n=7$) \\ \hline
1 & 11.2 (6.8) & 9.5 (4.5) \\ \hline
3 & 6.1 (1.6)  & 8.0 (1.5) \\ \hline
\end{tabular}
\end{table}

\section{Discussion and Conclusion}
In this study, we propose an evaluation method for robotic hand design using mechanical gloves. We also designed a three-DOF mechanical glove, conducted an evaluation experiment by a third party, and compared the performance of the three- and one-DOF configurations. The results show that there are no significant differences in the three- and one-DOF configurations on the completion time for almost all the object operations that are designed based on fine motor skills up to the age of four.

In general, it is important to control the position and orientation of a manipulating object. In this experiment, the movement of the arm was mechanically restricted to the six-DOF, similar to a general robotic arm; therefore, it is expected that the multiple DOF of the hand will contribute to the performance in many tasks. However, the results showed that the DOF did not significantly contribute to the performance although some tests required changes in the posture and orientation of the objects. In contrast, in the coin-grasping task, multiple DOFs significantly contributed to the performance. Grasping thin objects lying on flat and hard surfaces with a robotic hand is challenging \cite{Babin2019}. The results indicate that adding a few active joints improves the effectiveness in grasping of a thin object.

To generalize these results, similar experiments should be conducted using other hardware designs because of the following limitations. We cannot rule out the possibility that mechanical properties (such as the number of fingers, shape of the fingertip, and arrangement of the joints) and their parameters may change the effectiveness of the DOF. Furthermore, deprivation of sensory feedback owing to the rigid structure may affect the skill of both conditions. In addition, different skill levels may require different physical characteristics. These factors may yield different results. Conducting similar experiments under various conditions may reveal better tasks that exhibit clear differences in performance owing to the physical properties. The sophistication of the tasks will enable the efficient evaluation of mechanical designs. 

Many robotic designs need to be evaluated by independent institutions to reveal the contribution of embodiments to dexterity. We hope that the meta-analysis of these results will deepen the knowledge of embodiments. 




\section*{Appendix}
In this section, we provide the details of the evaluation tasks used in this study. All the tasks were designed to be completed within 45 s. If the working time exceeded 45 s, we considered it a failure and stopped the task. The details are as follows.

\begin{itemize}
    \item[1)] \textbf{Blocks}: Using only the right mechanical glove, the subject must build a stable tower using three toy blocks. The shapes and colors were green triangle, blue diamond, and purple ellipse. The initial postures were randomized, and we did not specify the order or direction of the blocks. Fences were set to enclose the working space to prevent objects from falling off the desk.
    
    \item[2)]\textbf{Jar}: The subject must open a jar using both hands. The completion time was recorded when the lid completely landed on the desk. The clamping pressure was controlled by drawing a line on the bottle.
    
    \item[3,4)]\textbf{Shape1 \& Shape2}: Using only the right mechanical glove, the subject must insert three blocks in the right place. The shapes are presented in the main text. The board was placed at the center of the boxed area, and blocks of randomized postures were placed on the right side of the area. The completion time was determined when the blocks were inserted onto the board, and fences were set around the working area.
    
    \item[5)]\textbf{Coin}: Using only the right mechanical glove, the subject must insert a coin in a piggy bank. Two coins were placed along the taped line, and a piggy bank was placed on the left of the coins. The completion time was determined when either coin fell in the piggy bank. 
    
    \item[6)]\textbf{Sticker}: Using both hands, the subject must put a sticker on a paper. The trial was finished when the subjects informed the experimenter about the completion or when the time limit was reached. The sticker sheet had four stickers at the beginning of the three trials and was used during the series of three trials.  
    
    \item[7)]\textbf{Page}: Using only the right mechanical glove, the subject must turn a single page of a loose-leaf file using a new sheet for each trial to control the condition of the paper. The completion time was determined when only a single page was placed on the right side of the filing ring.
    
    \item[8)]\textbf{Unbutton}: Using both hands, the subject must unbutton the clothes. The completion time was determined when achievement was observed by the experimenter.
    
\end{itemize}

\section*{Acknowledgement}
This study is partly based on the results obtained under a Grant-in-Aid for Scientific Research (A) JP18H04108 and partly on the results obtained from a project commissioned by the New Energy and Industrial Technology Development Organization (NEDO). 

\footnotesize
\bibliographystyle{IEEEtran}

\bibliography{biblio}

\end{document}